%% file: main.tex
\documentclass{llncs}

\usepackage[utf8]{inputenc}
\usepackage{amsmath}
\usepackage{graphicx}
\usepackage{amsfonts}
\usepackage{amssymb}
\usepackage{url}
\usepackage{caption}

\usepackage{hyperref}

\providecommand{\keywords}[1]{\textbf{Keywords: } #1}

\begin{document}

{\let\thefootnote\relax\footnotetext{Copyright \textcopyright\ 2020 for this paper by its authors. Use permitted under Creative Commons License Attribution 4.0 International (CC BY 4.0). CLEF 2020, 22-25 September 2020, Thessaloniki, Greece.}}

\title{Check\_square at CheckThat! 2020: \\Claim Detection in Social Media via Fusion of Transformer and Syntactic Features}

\author{Gullal S. Cheema\inst{1} \and Sherzod Hakimov\inst{1} \and Ralph Ewerth\inst{1,2}}
\institute{
TIB Leibniz Information Centre for Science and Technology, Hannover, Germany\\
\and
L3S Research Center, Leibniz University Hannover, Germany\\
\url{{gullal.cheema, sherzod.hakimov, ralph.ewerth}@tib.eu}}

\maketitle

\begin{abstract}
In this digital age of news consumption, a news reader has the ability to react, express and share opinions with others in a highly interactive and fast manner. As a consequence, fake news has made its way into our daily life because of very limited capacity to verify news on the Internet by large companies as well as individuals. In this paper, we focus on solving two problems which are part of the fact-checking ecosystem that can help to automate fact-checking of claims in an ever increasing stream of content on social media. For the first problem, claim check-worthiness prediction, we explore the fusion of syntactic features and deep transformer Bidirectional Encoder Representations from Transformers (BERT) embeddings, to classify check-worthiness of a tweet, i.e. whether it includes a claim or not. We conduct a detailed feature analysis and present our best performing models for English and Arabic tweets. For the second problem, claim retrieval, we explore the pre-trained embeddings from a Siamese network transformer model (sentence-transformers) specifically trained for semantic textual similarity, and perform KD-search to retrieve verified claims with respect to a query tweet. 
\end{abstract}


\input{sections/intro}

\input{sections/related_work}

\input{sections/proposed}

\input{sections/experiments}

\input{sections/conclusion}	

\section*{Acknowledgements}

This project has received funding from the European Union’s Horizon 2020 research and innovation programme under the Marie Skłodowska-Curie grant agreement no 812997.

\bibliographystyle{splncs04}
\bibliography{main}

\end{document}

%% file: sections/intro.tex
\section{Introduction}
Social media is increasingly becoming the main source of news for so many people. With around 2.5 billion Internet users, 12\% receive breaking news from Twitter instead of traditional media according to a 2018 survey report \cite{socialnews2018}. Fake news in general can be defined \cite{tandoc2018defining} as fabrication and manipulation of information and facts with the main intention of deceiving the reader. As a result, fake news can have several undesired and negative consequences. For example, recent news around COVID-19 pandemic with non-verified claims, that masks lead to rise in carbon dioxide levels caused an online movement to not wear masks. With ease of access and sharing news on Twitter, any news spreads much faster from the moment an event occurs in any part of the world. Although, the survey report \cite{socialnews2018} found that almost 60\% of users expect news on social media to be inaccurate, it still leaves millions of people who will spread fake news expecting it to be true. 

Considering the vast amount of news that spreads everyday, there has been a rise in independent fact-checking projects like \textit{Snopes, Alt News, Our.News}, who investigate the news that spread online and publish the results for public use. Most of these independent projects rely on manual efforts that are time consuming which makes it harder to keep up with rate of news production. Therefore, it has become very important to develop tools that can process news at a rapid rate and provide news consumers with some kind of an authenticity measure that reflects the correctness of claims in the news. 

In this paper, we focus on two sub-problems in CheckThat! 2020 \cite{clef-checkthat-en:2020}\footnote{\url{https://sites.google.com/view/clef2020-checkthat/}} that are a part of larger fact-checking ecosystem. In the first task, we focus on learning a model that can recognize check-worthy claims on Twitter. We present a solution that works for both English \cite{clef-checkthat-en:2020} and Arabic \cite{clef-checkthat-ar:2020} tweets. Some examples of tweets with claims are classified whether it is a check-worthy or not, shown in Table~\ref{tab:samp_t1}. One can see that the claims which are not check-worthy look like personal opinions and do not pose any threat to a larger audience. We explore the fusion of syntactic features and deep transformer Bidirectional Encoder Representations from Transformers (BERT) embeddings, to classify check-worthiness of a tweet. We use Part-of-speech (POS) tags, named entities, and dependency relations as syntactic features and a combination of hidden layers in BERT to compute tweet embedding. Before learning the model with a Support Vector Machine (SVM) \cite{suykens1999least}, we use Principal Component Analysis (PCA) \cite{wold1987principal} for dimensionality reduction. In the second task, we focus on learning a model that can accurately retrieve verified claims w.r.t a query claim, where query claim is a tweet and verified claims are snippets from actual documents. The verified claim is true and thus acts as the evidence or support for the query tweet. Some example pairs of tweets and claims can be seen in Table~\ref{tab:samp_t2}, which shows that the pairs share lots of contextual information which makes this task a semantic textual similarity problem. For this reason, we explore the pre-trained embeddings from a Siamese network transformer model (sentence-transformers) specifically trained for semantic textual similarity and perform KD-search to retrieve claims. We share the source code for both tasks publicly with the community. \footnote{\url{https://github.com/cleopatra-itn/claim_detection}}

The remainder of the paper is organized as follows. Section 2 briefly discusses about previous works on fake news detection and CheckThat! tasks in particular. Section 3 presents details of our approach for Task-1 and Task-2. Section 4 describes the experimental details and results. Section 5 summarizes our conclusion and future research directions. 

\begin{table}[!ht]
\caption{Sample tweets for Task-1 Check-Worthiness Prediction}
\label{tab:samp_t1}
	\centering
	\begin{tabular}{|l|c|c|}
		\hline
		\multicolumn{1}{|c|}{\textbf{Tweet}}                                                                                                                                                                                                & \textbf{Claim} & \textbf{Check-Worthy} \\ \hline
		\begin{tabular}[c]{@{}l@{}}Dear @VP Pence: What are you hiding from the \\ American people? Why won’t you let the people \\ see and hear what experts are saying about the \\ \#CoronaOutbreak?\end{tabular}                        & 0              & 0                     \\ \hline
		\begin{tabular}[c]{@{}l@{}}Greeting my good friends from the \#US the \\ \#Taiwan way. Remember: to better prevent the \\ spread of \#COVID19, say no to a handshake \& \\ yes to this friendly gesture. Check it out:\end{tabular} & 0              & 0                     \\ \hline
		\begin{tabular}[c]{@{}l@{}}Corona got these flights cheap as hell I gotta job \\ interview in Greece Monday \end{tabular}             & 1              & 0                     \\ \hline
		\begin{tabular}[c]{@{}l@{}}My mum has a PhD on Corona Virus from \\ WhatsApp University\end{tabular}                                                                                                                                & 1              & 0                     \\ \hline
		\begin{tabular}[c]{@{}l@{}}This is why the beaches haven't closed in \\ Florida, and why they've had minimal COVID-19 \\ prevention. Absolute dysfunction. \textless{}link\textgreater{}\end{tabular}                               & 1              & 1                     \\ \hline
		\begin{tabular}[c]{@{}l@{}}COVID-19 cases in the Philippines jumped \\ from 24 to 35 in less than 12 hours. This is \\ seriously ALARMING. Stay safe everyone! \\ \textless{}link\textgreater{}\end{tabular}                        & 1              & 1                     \\ \hline
	\end{tabular}
\end{table}

\begin{table}[!ht]
\caption{Sample pairs of tweets and verified claims for Task-2 Claim Retrieval}
	\label{tab:samp_t2}
	\centering
	\begin{tabular}{|c|l|}
		\hline
		\textbf{Tweet}          & \begin{tabular}[c]{@{}l@{}}A number of fraudulent text messages informing individuals \\ they have been selected for a military draft have circulated \\ throughout the country this week.\end{tabular}                                                                                                                \\ \hline
		\textbf{Verified Claim} & \begin{tabular}[c]{@{}l@{}}The U.S. Army is sending text messages informing people \\ they've been selected for the military draft.\end{tabular}                                                                                                                                                                       \\ \hline \hline
		\textbf{Tweet}          & \begin{tabular}[c]{@{}l@{}}El Paso was NEVER one of the MOST dangerous cities in \\ the US. We‘ve had a fence for 10 years and it has impacted \\ illegal immigration and curbed criminal activity. It is NOT \\ the sole deterrent. Law enforcement in our community \\ continues to keep us safe \#SOTU\end{tabular} \\ \hline
		\textbf{Verified Claim} & \begin{tabular}[c]{@{}l@{}}El Paso was one of the U.S. most dangerous cities before \\ a border fence was built there.\end{tabular}                                                                                                                                                                                    \\ \hline \hline
		\textbf{Tweet}          & \begin{tabular}[c]{@{}l@{}}Hey @Always since today is \#TransVisibilityDay it’s \\ probably important to point out the fact that this new \\ packaging isn’t trans* friendly. Just a reminder that \\ Menstruation does not equal Female. Maybe rethink \\ this new look. \textless{}link\textgreater{}\end{tabular}   \\ \hline
		\textbf{Verified Claim} & \begin{tabular}[c]{@{}l@{}}"In 2019, trans activists or ""the transgender lobby"" \\ forced Procter \& Gamble to remove the Venus symbol \\ from menstruation products."\end{tabular}                                                                                                                                  \\ \hline
	\end{tabular}
\end{table}

%% file: sections/related_work.tex
\section{Related Work}

Fake news has been studied from different perspectives in the last five years, like factuality or credibility detection \cite{popat2016credibility,giachanou2019leveraging,samadi2016claimeval,hassan2015detecting,hassan2017toward}, rumour detection \cite{zubiaga2017exploiting,zubiaga2018detection,sicilia2018twitter,zhao2015enquiring}, propagation in networks \cite{liu2018early,monti2019fake,shu2019studying,papanastasiou2020fake}, use of multiple modalities \cite{khattar2019mvae,wang2018eann,singhal2019spotfake} and also as an ecosystem of smaller sub-problems like in CheckThat! \cite{nakov2018overview,elsayed2019overview,clef-checkthat-lncs:2020}. For social media in particular, Shu \textit{et al.} \cite{shu2017fake} studied and provided a comprehensive review of fake news detection with characterizations from psychology and social science, and existing computational algorithms from data mining perspective. The fact that Twitter has become a source of news for so many people, researchers have extensively used the platform to formulate problems, extract data and test their algorithms. For instance, Zubiaga \textit{et. al.} \cite{zubiaga2017exploiting} extracted tweets around breaking news and used Conditional Random Fields to exploit context during the sequential learning process for rumour detection. Buntain \textit{et. al.} \cite{buntain2017automatically} studied three large Twitter datasets and developed models to predict accuracy assessments of fake news by crowd-sourced workers and journalists. While many approaches rely on tweet content for detecting fake news, there has been a rise in methods that exploit user characteristics and metadata to model the problem as fake news propagation. For example, Liu \textit{et. al.} \cite{liu2018early} modeled the propagation path of each news story as a multivariate time series over users who engaged in spreading the news via tweets. They further classified the fake news using Gated Recurrent Unit (GRU) and Convolutional Neural Networks (CNN) to capture the global and local variations of user characteristics respectively. Monti \textit{et. al.} \cite{monti2019fake} went a step further and used a hybrid feature set including user characteristics, social network structure and tweet content. They modeled the problem as binary prediction using a Graph CNN resulting in a highly accurate fake news detector.

Besides fake news detection, a sub task to predict check-worthiness of claims has also been explored recently mostly in political context. For example, Hassan \textit{et. al.} \cite{hassan2015detecting,hassan2016comparing} proposed a system that predicts the check-worthiness of a statement made by presidential candidates using SVM \cite{suykens1999least} classifier and combination of lexical and syntactic features. They also compared their results with fact-checking organizations like CNN\footnote{\url{http://www.cnn.com}} and PolitiFact\footnote{\url{https://www.politifact.com/}}. Later, in CheckThat! 2018 \cite{nakov2018overview}, several methods were proposed to improve the check-worthiness of claims in political debates. Best methods used a combination of lexical and syntactic features like Bag of Words (BoW), Parts-of-Speech (POS) tags, named Entities, sentiment, topic modeling, dependency parse trees and word embeddings \cite{mikolov2013distributed}. Various classifiers were built using either Recurrent Neural Networks (RNN) \cite{hansen2018copenhagen,zuo2018hybrid}, gradient boosting \cite{yasser2018bigir}, k-nearest neighbor \cite{ghanem2018upv} or SVM \cite{zuo2018hybrid}. In 2019 edition of CheckThat! \cite{elsayed2019overview}, in addition to using lexical and syntactic features \cite{gkasior2019ipipan}, top approaches relied on learning richer content embeddings and utilized external data for better performance. For example, Hansen \textit{et. al.} \cite{hansen2019neural} used word embeddings and syntactic dependency features as input to an LSTM network, enriched the dataset with additional samples from Claimbuster system \cite{hassan2017claimbuster} and trained the network with a contrastive ranking loss. Favano \textit{et. al.} \cite{favano2019theearthisflat} trained a neural network with Standard Universal Sentence Encoder (SUSE) \cite{cer2018universal} embeddings of the current sentence and previous two sentences as context. Another approach by Su \textit{et. al.} \cite{su2019entity} used co-reference resolution to replace pronouns with named entities to get a feature representation with bag of words, named entity similarity and relatedness. Other than political debates, Jaradat \textit{et. al.} \cite{jaradat2018claimrank} proposed an online multilingual check-worthiness system that works for different sources (debates, news articles, interviews) in English and Arabic . They use actual annotated data from reputable fact-checking organizations and use best performing feature representations from previous approaches. For tweets in particular, Majithia \textit{et. al.} \cite{majithia2019claimportal} proposed a system to monitor, search and analyze factual claims in political tweets with Claimbuster \cite{hassan2017claimbuster} at the backend for check-worthiness. Lastly, Dogan \textit{et. al.} \cite{dogan2015detecting} also conducted a detailed study on detecting check-worthy tweets in U.S. politics and proposed a real-time system to filter them.

%% file: sections/proposed.tex
\section{Proposed Approach}
\subsection{Task-1: Tweet Check-Worthiness Prediction}
Check-Worthiness prediction is the task of predicting whether a tweet includes a claim that is of interest to a large audience. Our approach is motivated by the successful use of lexical, syntactic and contextual features in the previous editions of CheckThat! check-worthiness task for political debates. Given that this task contains less amount of training data, we approached this problem with the idea of creating a rich feature representation, reducing the dimensions of large feature set with PCA \cite{wold1987principal} and then learning the model with a SVM. In doing so, our goal is also to understand which features are the most important for check-worthiness prediction from tweet content. As context is very important for downstream NLP tasks, we experiment with word embeddings (word2vec \cite{mikolov2013distributed}, GloVe \cite{pennington2014glove})  and BERT \cite{devlin2018bert} embeddings to create a sentence representation of each tweet. Our pre-processing and feature extraction is agnostic to the topic of the tweet so that it can be applied to any domain. Next, we provide details about all the features used, their extraction and the encoding process. Our overall approach can be seen in Figure~\ref{fig:app_task1}.

\subsubsection{Pre-processing}
We use two publicly available pre-processing tools for English and Arabic tweets. We use Baziotis \textit{et. al.}'s \cite{baziotis-pelekis-doulkeridis:2017:SemEval2} tool for English to apply the following normalization steps: tokenization, lower-casing, removal of punctuation, spell correction, normalize \textit{hashtags, all-caps, censored, elongated and repeated} words, and terms like \textit{URL, email, phone, user mentions}. We use Stanford Stanza \cite{qi2020stanza} toolkit to pre-process Arabic tweets by applying the following normalization steps: tokenization, multi-word token expansion and lemmatization.

In the case of extracting word embeddings from a transformer network, we use the raw text as the networks have their own tokenization process. 

\subsubsection{Syntactic Features}

We use the following syntactic features for English and Arabic tasks: Parts-of-Speech (POS) tags, named entities (NE) and dependency parse tree relations. We use the pre-processed text and run off-the-shelf tools to extract syntactic information of tweets and then convert each group of information to feature sets. For English we used spaCy\cite{honnibal2017spacy} and Stanford Stanza \cite{qi2020stanza} for Arabic tweets to extract the following syntactic features. In all the features, we experiment with keeping and removing stop-words to evaluate their affect.

\textbf{Part-of-Speech}: For both English and Arabic, we extract 16 POS tags in total and through our empirical evaluation we find that the following eight tags to be the most useful when used as features: NOUN, VERB, PROPN, ADJ, ADV, NUM, ADP, PRON. For Arabic, the additional four tags are useful features: DET, INTJ, AUX, PART. We used the chosen set of POS tags for respective language to encode the syntactic information of tweets.

\textbf{Named Entities}: We identified the following named entity types to be the most important features through our evaluation: (GPE, PERSON, ORG, NORP, LOC, DATE, CARDINAL, TIME, ORDINAL, FAC, MONEY) for English and (LOC, PER, ORG, MISC) for Arabic. We also found that while developing feature combinations named entities do not add much value to overall accuracy, and hence our primary and contrastive submissions do not include them.

\begin{figure}[ht]
	\centering
	\includegraphics[width=\linewidth]{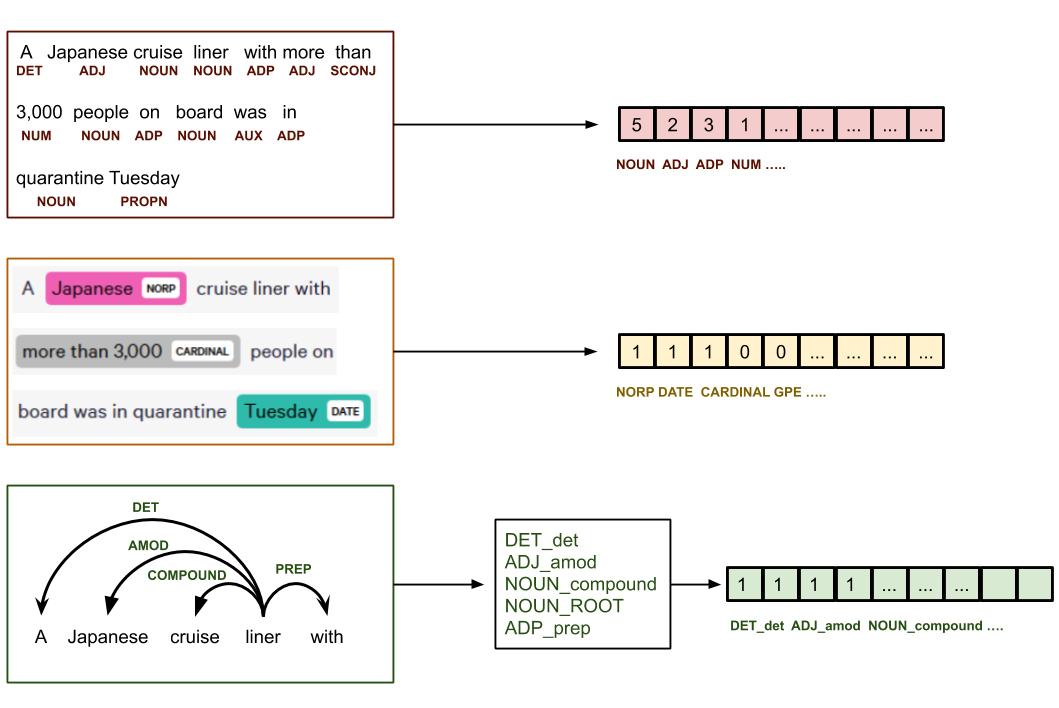}
	\caption{Syntactic feature extraction and encoding process. Feature vectors are based on the number of times it is seen in the given sentence.}
	\label{fig:feat_extract}
	\vspace{-0.2cm}
\end{figure}

\textbf{Syntactic Dependencies}: these features are constructed using dependency relation between tokens in a given tweet. We use the dependency relation between two nodes in the parsed tree if the the child and parent nodes' POS tags are one of the following ADJ, ADV, NOUN, PROPN, VERB or NUM. All dependency relations that match the defined constraint are converted into the triplet relation such as (\emph{child node-POS, dependency-relation, parent-POS}) and pairs such as (\emph{child node-POS, dependency-relation}) where the relation is not part of a feature representation. This process is shown in Figure~\ref{fig:feat_extract}. We found that the features based on pairs of child and parent node perform better than the triplet feature. The dimension of the feature vector for English and Arabic is 133 and 134 respectively.

For encoding a feature, we get a histogram vector which contains the number of type of tag, named entity or syntactic relation pair. The process of feature encoding is shown in Figure~\ref{fig:feat_extract}. Finally, we normalize each type of feature with maximum value in the vector.

\subsubsection{Average Word Embeddings}
One simple way to get a contextual representation of a sentence is to average the word embeddings of each token in a given sentence. For this purpose, we experiment with three types of word embeddings pre-trained on three different sources for English: GloVe embeddings  \cite{pennington2014glove} trained on Twitter and Wikipedia, word2vec embeddings \cite{mikolov2013distributed} trained on Google News, and FastText \cite{mikolov2017advances} embeddings trained on multiple sources. In addition, we also experiment with removing stop-words from the average word representation, as stop-words can dominate in the average and result in less meaningful sentence representation. For Arabic, we use word2vec embeddings that are trained on Arabic tweets and Arabic Wikipedia \cite{soliman2017aravec}.

\subsubsection{Transformer Features}
Another way to extract contextual features is to use BERT \cite{devlin2018bert} embeddings that are trained using the context of the word in a sentence. BERT is usually trained on a very large text corpus which makes them very useful for off-the-shelf feature extraction and fine-tuning for downstream tasks in NLP. To get one embedding per tweet, we follow the observations made in \cite{devlin2018bert} that, different layers of BERT capture different kinds of information, so an appropriate pooling strategy should be applied depending on the task. The paper also suggests that the last four hidden layers of the network are good for transfer learning tasks and thus we experiment with 4 different combinations, i.e. concatenate last 4 hidden layers, average of last 4 hidden layers, last hidden layer and $2^{nd}$ last hidden layer. We normalize the final embedding so that $l2$ norm of the vector is 1. We also experimented with BERT's pooled sentence embedding that is encoded in the \emph{CLS} (class) tag, which performed significantly poorer than the pooling strategies we employed. For Arabic, we only experimented with a sentence-transformer \cite{reimers2019sentence} that is trained on multilingual training corpus and outputs a sentence embedding for each tweet/sentence.

\begin{figure}[t]
	\centering
	\includegraphics[width=\linewidth]{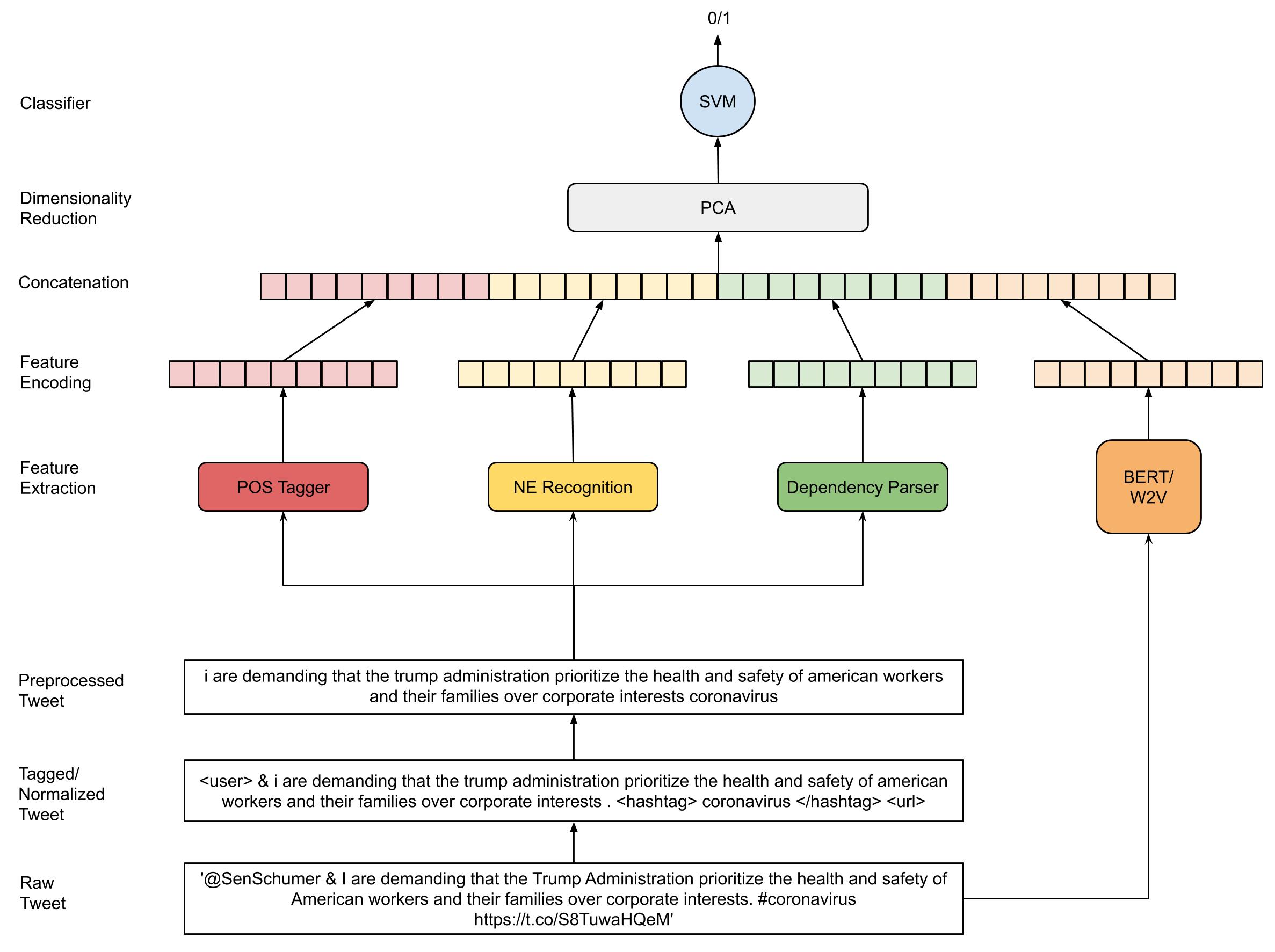}
	\caption{Proposed Approach for Check-Worthiness Prediction}
	\label{fig:app_task1}
\end{figure}

\textbf{Sentence Representation}: To get the overall representation of the tweet, we concatenate all the syntactic features together with either average word embedding or BERT-based transformer features and then apply PCA for dimensionality reduction. SVM classifier is trained on the feature vectors of tweets to output a binary decision (check worthy or not). The overall process is shown in Figure~\ref{fig:app_task1}.

\subsection{Task-2: Claim Retrieval}
Claim Retrieval is the task of retrieving the most similar already verified claim to the query claim. For this task, it is important that the feature representation captures the meaning and context of words and phrases so that query matches the correct verified claim. Therefore, we relied on a triplet-network setting, where the network could be trained with triplets consisting of an anchor sample \textit{a}, positive sample \textit{p} and a negative sample \textit{n}. We use triplet loss to fine-tune a pre-trained sentence embedding network, such that the distance between \textit{a} and \textit{p} is smaller than the distance between \textit{a} and \textit{n} using the following loss function. 

\begin{align}
Loss = \sum_{i=1}^{N} [||S_i^a - S_i^p||_2^2 - ||S_i^a - S_i^n||_2^2 + m]_+
\end{align}


where $S_i^a$, $S_i^p$ and $S_i^n$ are triplet sentence embeddings and $m$ is the margin (set to 1), \textit{N} is the number of samples in the batch.

As each verified claim is a tuple consisting of text and title, we create two triplets for every true tweet-claim pair, i.e., (anchor tweet, true claim text, negative claim text) and (anchor tweet, true claim title, negative claim title). This increases the number of positive samples for training as there are only 800 samples and one true claim for every tweet. To get negative claims, we select 3 claims with highest cosine similarity that are not the true claims for the anchor tweet using the pre-trained sentence-transformer embeddings. For pre-processing, we use Baziotis \textit{et. al.}'s \cite{baziotis-pelekis-doulkeridis:2017:SemEval2} tool for processing the tweets to remove \textit{URL, email, phone, user mentions}, as the claim text or title do not contain any such information.

As retrieval is a search task, we used KD-Tree search to find the most similar already verified claim that has the minimum Euclidean distance to the query claim. The sentence embeddings extracted from the network are used to build a KD-Tree and for each query claim, top 1000 verified claims are extracted from the tree for evaluation. For building the KD-Tree, we average the sentence embeddings of claim text and claim title, as it performs better than just using either claim or title. In our ablation study, we directly compute the cosine similarity between each query tweet and all the verified claims, and pick the top 1000 (highest cosine similarity) verified claims for evaluation. We conduct the second evaluation because building a KD-Tree can affect the retrieval accuracy.

\subsubsection{Sentence Transformers for Textual Similarity}
As a backbone network to extract sentence embeddings and fine-tuning with triplet loss, we use the recently proposed Sentence-BERT \cite{devlin2018bert} that learns the embeddings in a Siamese (pairs) and triplet network settings. We experiment with the pre-trained Siamese Network models trained on SNLI (Stanford Natural Language Inference) \cite{bowman2015large} and STSb (Semantic Textual Similarity benchmark) \cite{cer2017semeval} datasets that have been shown to perform very well for semantic textual similarity.

%% file: sections/experiments.tex
\section{Experiments and Results}

\subsection{Task-1: Tweet Check-Worthiness Prediction}
\subsubsection{Dataset and Training Details}
English dataset consists of training, development (dev) and test splits with 672, 150 and 140 tweets respectively on the topic of COVID-19. We perform grid search using development set to find the best parameters. Arabic dataset consists of training and test splits with 1500 tweets on 3 topics and 6000 tweets on 12 topics respectively with 500 tweets on each topic. For validation purpose, we keep 10\% (150 samples) from the training data as development set. The official ranking of submitted system for this task is based on Mean Average Precision (MAP) and Precision@30 (P@30) for English and Arabic datasets, respectively.

To train the SVM models for both English and Arabic, we perform grid search over PCA energy (\%) conservation, regularization parameter \textit{C} and RBF kernel's \textit{gamma}. Parameters range for PCA varies from 100\% (original features) to 95\% with decrements of 1, and both \textit{C} and \textit{gamma} vary between -3 to 3 on a log-scale with 30 steps. For faster training on a large grid search, we use ThunderSVM \cite{wenthundersvm18} which takes advantage of a GPU or a multi-core system to speed up SVM training. 

\subsubsection{Results}
Our submissions used the best models that we obtained from the grid search and are briefly discussed below.

\textbf{English}: We made 3 submissions in total. Our primary (Run-1) and $2^{nd}$ contrastive (Run-3) submission uses sentence embeddings computed from BERT-large word embeddings as discussed in the proposed work section. In addition, both submissions use POS tag and dependency relation features. Interestingly, we found that the best performing sentence embeddings did not include stop-words. The primary submission (Run-1) uses an ensemble of predictions from three models trained on concatenated last 4 hidden layers, average of last 4 hidden layers and $2{nd}$ last hidden layer. The $2^{nd}$ contrastive submission (Run-3) uses predictions from the model trained on the best performing sentence embedding computed from concatenating last 4 hidden layers. Our $1^{st}$ contrastive submission (Run-2) uses an ensemble of predictions from three models trained with GloVe\cite{pennington2014glove} on Twitter with 25, 50 and 100-dimensional embeddings but with the same POS tag and dependency relation features. We use majority voting to get the final prediction and mean of decision values to get the final decision value. We found that removing the stop-words to compute average of word embeddings actually degraded the performance and hence included them in the average. 

We also add some additional results to see the effect of stop-words, POS tags, named entities, dependency relations and ensemble predictions in Table~\ref{tab:eng_t1}. The effect of stop-words can be clearly seen in alternative runs of Run-1 and Run-3, where the MAP clearly drops by 1-2 points. Similarly, the negative effect of removing POS tag and dependency relation features can be seen in rest of the alternative runs. Lastly, adding named entity features to the original submissions also decreases the precision by 1-2 points. This might be because the tweets have very few named entities and are not useful to distinguish between check-worthy and not check-worthy claims. For comparison with other teams in the challenge, we show top 3 results at the bottom of the table for reference. Team Accenture \cite{clef-checkthat-williams:2020} fine-tuned a RoBERTa model with an extra mean pooling and a dropout layer to prevent overfitting. Team Alex \cite{clef-checkthat-Nikolov:2020} experimented with different tweet pre-processing techniques and various transformer models together with logistic regression and SVM. Their main submission used logistic regression trained on 5-fold predictions from RoBERTa concatenated with tweet metadata. Team QMUL-SDS \cite{clef-checkthat-Alkhalifa:2020} fine-tuned a BERT model pre-trained specifically on COVID twitter data.

\bgroup
\setlength{\tabcolsep}{5pt} 
\renewcommand{\arraystretch}{1.5} 
\begin{table}[ht]
\caption{Task-1 Check-Worthiness English Results, MAP (Mean Average Precision),  DRel (dependency relations), NE (named entities), *Primary Submission}
	\label{tab:eng_t1}
	\centering
	\begin{tabular}{|l|cccccc|c|}
		\hline
		\textbf{Run} & \textbf{Stopwords} & \textbf{Ensemble} & \textbf{POS} & \textbf{DRel} & \textbf{NE} & \textbf{Embedding} & \textbf{MAP} \\ \hline
		Run-$1^*$        &                    &   \checkmark           & \checkmark  & \checkmark &    & BERT   & \textbf{0.7217}       \\ \hline
		Run-2        &        \checkmark            &       \checkmark      & \checkmark  &   \checkmark &  & GloVe  & 0.6249       \\ \hline
		Run-3        &                    &               & \checkmark  &  \checkmark &  & BERT & 0.7139      \\ \hline \hline
		Run-1-1      &      \checkmark              &   \checkmark         &  \checkmark  & \checkmark &   & BERT    & 0.7102              \\ \hline
		Run-1-2      &                    &   \checkmark         & \checkmark    & &    & BERT     & 0.6965              \\ \hline
		Run-1-3      &                    &   \checkmark         &    & &   & BERT    &  0.7094             \\ \hline
		Run-1-4      &                    &   \checkmark           & \checkmark  & \checkmark & \checkmark & BERT    &  0.7100           \\ \hline
		Run-3-1      &      \checkmark              &               & \checkmark  & \checkmark &  & BERT  & 0.6889              \\ \hline
		Run-3-2      &                    &           & \checkmark     & &    & BERT    &  0.7074             \\ \hline
		Run-3-3      &                    &          &    &  &   & BERT    &    0.6981           \\ \hline
		Run-3-4      &                    &               & \checkmark  &  \checkmark & \checkmark & BERT   &  0.6940            \\ \hline
		\hline
		\cite{clef-checkthat-williams:2020} & - & - & - & - & - & - & 0.8064 \\ \hline
		\cite{clef-checkthat-Nikolov:2020} & - & - & - & - & - & - & 0.8034 \\ \hline
		\cite{clef-checkthat-Alkhalifa:2020} & - & - & - & - & - & - &  0.7141 \\ \hline
	\end{tabular}
\end{table}
\egroup

\textbf{Arabic} There are a total of four submissions that we made in this task. Our best performing submission (Run-1) uses 100-dimensional word2vec Arabic embeddings trained on a Twitter corpus \cite{soliman2017aravec} in combination with POS tag features. Our second and third submissions are redundant in terms of feature use, so we only mention the second one (Run-2) here. In addition features used in first submission, it uses dependency relation features and 300-dimensional Twitter embeddings instead of 100-dimensional. Our last submission (Run-3) uses only pre-trained multilingual sentence-transformer\footnote{\url{https://github.com/UKPLab/sentence-transformers}} \cite{reimers-2020-multilingual-sentence-bert}  that is trained on 10 languages including Arabic. In the first three submissions, we removed the stop-words from all the features as keeping them resulted in a poorer performance.  \textit{Precision@K} and \textit{Average Precision} (AP) results on the test set are shown in the same order in Table~\ref{tab:arab_t1}. Official metric for ranking is P@30. For comparison with other teams in the challenge,we show top 3 results at the bottom of the table for reference. Team Accenture \cite{clef-checkthat-williams:2020} experimented with and fine-tuned three different pre-trained Arabic BERT models and used external data to increase the positive instances. Team TOBB-ETU \cite{clef-checkthat-Kartal:2020} used logistic regression and experimented with Arabic BERT and word embeddings together to classify tweets. Team UB\_ET \cite{clef-checkthat-Hasanain:2020} used a multilingual BERT for ranking tweets by check-worthiness.

\bgroup
\setlength{\tabcolsep}{6pt} 
\renewcommand{\arraystretch}{1.5} 
\begin{table}[htp]
\caption{Task-1 Check-Worthiness Arabic Results, P@K (Precision@K) and AP (Average Precision), *Primary Submission}
	\label{tab:arab_t1}
	\centering
	\begin{tabular}{|l|c|c|c|c|c|c|c|}
		\hline
		\textbf{Run ID} & \textbf{P@5} & \textbf{P@10} & \textbf{P@15} & \textbf{P@20} & \textbf{P@25} & \textbf{P@30} & \textbf{AP} \\ \hline
		Run-$1^*$           & \textbf{0.6000}       & \textbf{0.6083}        & \textbf{0.5944}        & \textbf{0.6000}        & \textbf{0.5900}        & \textbf{0.5778}        & \textbf{0.4949}      \\ \hline
		Run-2           & 0.5500       & 0.5667        & 0.5611        & 0.5417        & 0.5433        & 0.5361        & 0.4649      \\ \hline
		Run-3           & 0.4000       & 0.3917        & 0.4167        & 0.4292        & 0.4433        & 0.4472        & 0.4279      \\ \hline
		\hline
		\cite{clef-checkthat-williams:2020} & 0.7333 &	0.7167 &	0.7167 &	0.6875 &	0.6933 &	0.7000 &	0.6232 \\ \hline
		\cite{clef-checkthat-Kartal:2020} & 0.7000 &	0.7000 &	0.7000 &	0.6625 &	0.6500 &	0.6444 &	0.5816\\ \hline
		\cite{clef-checkthat-Hasanain:2020} & 0.6833 &	0.6417 &	0.6667 &	0.6333 &	0.6367 &	0.6417 &	0.5511 \\ \hline
	\end{tabular}
	
\end{table}
\egroup

\subsection{Task-2: Claim Retrieval}
\subsubsection{Dataset and Training Details}
The dataset in this task has 1,003 tweets for training and 200 tweets for testing. These tweets are to be matched against a set 10,373 verified claims. From the training set, 197 tweets are kept for validation. To fine-tune the sentence-transformer network with the triplet loss, we use a batch size of eight and train the network for two epochs. The official ranking of this is based on Mean Average Precision@5 (MAP@5). All tweets and verified claims are in English.

Our primary (Run-1) and $2^{nd}$ contrastive (Run-3) submission uses BERT-base and BERT-large pre-trained on SNLI dataset with sentence embedding pooled from the \emph{CLS} and \emph{MAX} tokens respectively.  We fine-tune these two networks with the triplet loss.
On the contrary, our $1^{st}$ contrastive submission (Run-2) uses multilingual DistilBERT model \cite{reimers-2020-multilingual-sentence-bert} trained on 10 languages including English. This model is directly used to test the pre-trained embeddings.

\subsubsection{Results}
Interestingly, pre-trained embeddings extracted from multilingual DistilBERT without any fine-tuning turn out to be better for semantic similarity than fine-tuned monolingual BERT models. Having said that, the fine-tuned monolingual BERT models do perform better than extracted pre-trained embeddings and the difference can be seen in \textit{Run-1-2} and \textit{Run-3-2} in Table~\ref{tab:eng_t2}. We also try to fine-tune the multilingual model which drastically decreases the retrieval performance. The decrease can be attributed to the pre-training process \cite{reimers-2020-multilingual-sentence-bert} in which the model was trained in a teacher-student knowledge distillation learning framework and on multiple languages. As stated in the proposed work section, we conduct a second evaluation to retrieve the claims with highest similarity without KD-Search and the results are significantly better as shown in Table~\ref{tab:eng_t2}. For comparison with other teams in the challenge, we have shown top 3 primary submissions at the bottom of the table for reference. Team Buster.AI \cite{clef-checkthat-Bouziane:2020} investigated sentence similarity using transformer models, and experimented with multimodality and data augmentation. Team UNIPI-NLE \cite{clef-checkthat-Passaro:2020} fine-tuned a sentence-BERT in two steps, first to predict the cosine similarity of positive and negative pairs, followed by a binary classification of whether a tweet-claim pair is a correct match or not. Team UB\_ET \cite{clef-checkthat-Thuma:2020} experimented with three different models to rank the verified claims and their main submission used a DPH Divergence from Randomness (DFR) term weighting model.

\bgroup
\setlength{\tabcolsep}{3pt} 
\renewcommand{\arraystretch}{1.5} 
\begin{table}[htp]
\caption{Task-2 Claim Retrieval Results, MAP (Mean Average Precision), *Primary Submission}
	\label{tab:eng_t2}
	\centering
	\begin{tabular}{|l|cc|c|c|c|c|}
		\hline
		\textbf{Run} & \textbf{Fine-tuned} & \textbf{KD-Search} & \textbf{MAP@1}  & \textbf{MAP@3}  & \textbf{MAP@5}  & \textbf{MAP@10} \\ \hline
		Run-$1^*$        &      \checkmark               &     \checkmark               & 0.6520          & 0.6900          & 0.6950          & 0.7000          \\ \hline
		Run-2        &                     &      \checkmark              & \textbf{0.8280} & \textbf{0.8680} & \textbf{0.8730} & \textbf{0.8740} \\ \hline
		Run-3        &      \checkmark               &      \checkmark              & 0.7180          & 0.7460          & 0.7540          & 0.7600          \\ \hline \hline
		Run-1-1      &      \checkmark               &                    & 0.703           & 0.743           & 0.756           & 0.760           \\ \hline
		Run-1-2      &                     &                    & 0.527           & 0.584          & 0.589           & 0.594           \\ \hline
		Run-2-1      &                     &                    & 0.858           & 0.892           & 0.894           & 0.896           \\ \hline
		Run-3-1      &      \checkmark               &                    & 0.718           & 0.764           & 0.770           & 0.772           \\ \hline
		Run-3-2      &                     &                    & 0.532           & 0.569           & 0.576           & 0.585           \\ \hline
		\hline
		\cite{clef-checkthat-Bouziane:2020} & - & - & 0.8970 &	0.9260 &	0.9290 &	0.9290  \\ \hline
		\cite{clef-checkthat-Passaro:2020} & - & - & 0.8770 & 	0.9070 &	0.9120 &	0.9130  \\ \hline
		\cite{clef-checkthat-Thuma:2020} & - & - & 0.8180 &	0.8620 &	0.8640 &	0.8660 \\ \hline
	\end{tabular}
	
\end{table}
\egroup

%% file: sections/conclusion.tex
\section{Conclusion and Future Work}
In this paper, we have presented our solutions for two tasks in CLEF CheckThat! 2020. In the first task, we used syntactic, contextual features and SVM for predicting the check-worthiness of tweets in Arabic and English. For syntactic features, we evaluated Parts-of-Speech tags, named entities and syntactic dependency relations, and used the best feature sets for both languages. In the case of contextual features, we evaluated different word embeddings, BERT models and sentence-transformers to capture the semantics of each tweet or sentence. For future work, we would like to evaluate the possibility of using relevant metadata and other modalities like images and videos present in tweets for claim's check-worthiness. In the second task, we evaluated monolingual and multilingual sentence-transformers to retrieve verified claims for the query tweet. We found that off-the-shelf multilingual sentence-transformer is very well suited for semantic textual similarity task than other monolingual BERT models.